\documentclass[conf]{new-aiaa}
\usepackage[utf8]{inputenc}

\usepackage{amsmath}
\usepackage{amssymb,amsfonts}
\usepackage{algorithmic}
\usepackage{graphicx}
\usepackage{textcomp}
\usepackage{xcolor}
\usepackage{amsmath}
\usepackage[version=4]{mhchem}
\usepackage{siunitx}
\usepackage{longtable,tabularx}
\usepackage{derivative}
\usepackage{physics}
\title{Path Planning Using Deep Deterministic Policy Gradient: a Reinforcement Learning Approach
\footnote{DISTRIBUTION STATEMENT A. Approved for public release: distribution is unlimited (AFRL-2026-2026-1882) Cleared 16 Apr 2026}
}

\author{Qiang Le\footnote{Associate Professor, Department of Electrical and Computer Engineering Hampton University, Hampton, VA}}
\affil{Hampton University, Hampton, VA 23669, USA}

\author{Yaguang Yang\footnote{Research Scientist, Department of Electrical and Computer Engineering Hampton University, Hampton, VA}}
\affil{Hampton University, Hampton, VA 23669, USA}

\author{Isaac E. Weintraub\footnote{Control Science Center, Air Force Research Laboratory, AIAA Associate Fellow. This paper is based on work performed at the Air Force Research Laboratory (AFRL) Control Science Center and is supported by AFOSR LRIR \#24RQCOR002 (funded by Dr. Frederick Leve).}}
\affil{Air Force Research Laboratory, Wright-Patterson AFB, OH 45433, USA}

\begin{document}

\maketitle

\begin{abstract}
Path-planning for autonomous vehicles in threat-laden environments is a fundamental challenge because the problem is nonlinear and nonconvex even in simplest scenarios. While traditional optimal control methods can be used to find ideal paths, the computational time is often too slow for real-time decision-making. To solve this challenge, we propose a method based on Deep Deterministic Policy Gradient (DDPG) and model the threat as possibly multiple circular 'no-go' zones. A mission is regarded as a failure if the vehicle enters this restricted zone at any time or does not reach a neighborhood of the destination. The DDPG agent is trained through trial and error in a simulated environment, learning a direct mapping from its current state (position and heading) to a series of feasible actions that guide the agent to safely reach its destination. The reword function has three parts: (a) an attractive field centered at the final destination, (b) some repulsive fields centered at the origins of circular obstacles, and (c) a penalty of control energy consumption (the magnitude of heading change) that indirectly in favor for straight path.
The DDPG trains the agent using these incentives to find a largest possible set of starting points wherein a safe path to the destination is guaranteed. This provides critical information for mission planning, showing beforehand whether a task is achievable from a given starting point, assisting pre-mission planning activities. The approach is validated in simulation. A comparison between the DDPG method and a traditional optimal control (pseudo-spectral) method is carried out. The results show that the learning-based agent produces effective paths while being significantly faster, making it a better fit for real-time applications. 
\end{abstract}


\section{Introduction}

Path planning in the presence of obstacles has many applications, for example, in engagement zone avoidance \cite{dzwv23,vw24,wvchf22}, robotics path planning \cite{gc13}, automatic driving \cite{hctch18}, video games \cite{ask15}, space debris avoidance \cite{msml11}, printed circuit board routing \cite{zwlw22}, geographical information system \cite{bg07}, and practical VLSI systems \cite{up23}, and many more. 
Different methods, such as optimal control (pseudo spectral and model predictive control method) \cite{herber15}, machine learning \cite{hwmz20}, Genetic Algorithm (GA), and Voronoi diagram \cite{nstj19}, have been developed to solve the path planning problem in the presence of obstacles. While the traditional optimal control method can find an optimal path, developing a high-fidelity model can be difficult and time-consuming. Moreover, the computational time is often too slow for real-time decision-making. Genetic Algorithm and Voronoi diagram may not need a high-fidelity model, but the computational time can still be a problem for real-time decision-making. Recently, reinforcement learning emerged as a popular method for path planning in the presence of obstacles because it does not require a high-fidelity model and a well-trained agent can find a feasible path in real-time in most scenarios.

Some popular reinforcement learning methods for path planning in the presence of obstacles include Q-learning, Deep Q-Network (DQN), and Deep Deterministic Policy Gradient (DDPG). Q-learning was first proposed by Watkins in 1989 in \cite{watkins89}. Many important reinforcement learning methods are based on or adopted from the ideas of Q-learning. Since Q-learning is naturally related to the dynamical programming, it has been used to solve path planning problem \cite{cx18,mh20,prppl24,syc23,wyl22,zzzy17}. One of the main drawbacks of the Q-learning method is that it can only handle discrete state space and action space problems which precludes it from solving large problems which have continuous state and action spaces. For this reason, Deep Q-Network method Q-network (DQN) was developed by a group of engineers at Google in 2015 \cite{mksgawr13,mnih15,mbmglhsk16} to handle problems with continuous state space. Clearly, DQN can be applied to more general path planning problems. As a matter of fact, DQN is quickly used for path planning problem in many publications, for example, \cite{cpll22,gzlsx23,hza25,hqll23,lzd18,nkm23,wzdlzll23,xyj25,ylp20,yzqygzly20,zlxg18}. Although DQN is able to handle continuous state space problems using a deep neural network, its application is limited to problems with discrete action space. To extend the Q-learning and DQN methods to handle problems with continuous state space and action space, Lillicrap et al \cite{lhphetsw15} proposed Deep Deterministic Policy Gradient method by using Deterministic Policy Gradient developed by Silver et al \cite{slhdwr14} and combining the ideas from Q-learning and DQN. Immediately, this method is applied to the path planning problems by Wang et al \cite{wcwzh18} and followed by many researchers \cite{asz24,akr23,gylwc23,haws20, rhv23,whlm23,xcayz22}. Many of these applications include modifications and improvement of DDPG. For example, Ali et al \cite{asz24} proposed a novel reward function which employs quaternions for orientation control to reduce pose misalignments and dynamic singularities, as opposed to traditional Euler angles; Gao et al \cite{gylwc23} combined DDPG with long short-term memory (LSTM) network-based encoder to achieve dynamic obstacle avoidance for the mobile robot in the stochastic working scenario; He \cite{haws20} et al combined imitation learning with reinforcement learning and building upon Twin Delayed DDPG (TD3) algorithm to speed up the deep reinforcement learning (DRL) training process; Ramezani et al \cite{rhv23} suggested to combine the model predictive control (MPC) with DDPG, aiming at improving the performance of DDPG and overcoming the computational burden of MPC; Wang et al \cite{whlm23} converted the obstacle avoidance problem to an optimal learning incentive problem, the obstacle avoidance policy was learnt from DDPG, and the obstacle avoidance trajectory evaluation function was optimized using the DDPG reward incentive mechanism; Finally, Xu et al \cite{xcayz22} used DDPG for path planning and dynamic collision avoidance algorithm in marine ship navigation.

In this paper, we combine the DDPG with ideas from different considerations to solve path following problem in the presence of obstacles. First an attractive artificial potential field is used to captivate the agent from any location to the desired destination. Second, repulsive artificial potential fields are used to prevent the agent from entering ``no-go'' zones. Third, minimum heading change enforces the agent spends least fuels, which indirectly minimizes the path length. 
Finally, a smart initial heading is calculated to prevent the agent from moving toward the ``no-go'' zones, which greatly increases the chance to avoid the obstacles in the early steps. These strategies significantly improve the performance of the trained agent that was obtained in \cite{lyw26}. The proposed method is implemented in Matlab.
To validate that DDPG works for more complicated situations, we extend the method to a path planning problem with three obstacles and demonstrate its effectiveness.

The remainder of the paper is organized as follows. Section \ref{DDPGmethod} discusses our method which is based on DDPG. Section \ref{singleObs} considers the application of DDPG to path planning in the presence of one obstacle, which provides more details about the training and re-training, effects of smart heading.
Section \ref{multiObs} investigates the path planning problem with multiple obstacles, which shows that the proposed DDPG method is still effective. The last section summarizes the conclusions and future work.

\section{Deep Reinforcement Learning}\label{DDPGmethod}

Reinforcement learning (RL) is model-free, trial-and-error-based machine learning where an agent learns to take actions in different situations in a given environment to maximize the long-term accumulative rewards. The Q-learning is the first Reinforcement Learning method. In Q-learning, the agent updates the Q-table that stores the expected discounted long-term accumulative reward for a given observation and action. Q-learning can find the optimal path given a starting position, the final position, and some static block-out obstacles while the action space of the agent is small, for example 8 directions: E, W, N, S, SW SE, NE, and NW. When the grid-based environment and the number of actions are growing big, the size of Q-table will explode and become impossible to update. This leads to the development of deep learning or neural networks to build DQN. Its main idea is to combine Q-learning with a neural network so that DQN can estimate the accumulative long time reward defined by Bellman equation. Since the action space for the DQN agent is still discrete, DDPG therefore introduced a neural actor network to tackle the continuous state space and action space.

\subsection{Actor and critic neural networks}

DDPG combines two neural networks: the actor network and critic network with Q-learning to explore continuous state space and action space \cite{lhphetsw15}.
Instead of using a probabilistic policy $\pi$ for an action $A=\pi(S;\theta_A)$ in DQN, DDPG represents the probabilistic policy by a deterministic policy
$A_{\theta}$ which is a map from S to A using the existing network parameters $\theta_A$ for the given state $S$: 
\begin{equation}
A_{\theta}=\mu(S;\theta_A),
\label{policy}
\end{equation}
The output of the critic network is the Q value for the given state $S$ and the action $A$ based on the critic network parameters $\theta_C$: 
\begin{equation}
Q_{A}(S,A; \theta_C)= E[R_1 ~|~ S, A_{\theta}; \theta_C],
\label{cNetwork}
\end{equation} 
where 
\begin{equation}
R_1=\sum_{i=1}^{T} \gamma^{i-t} r(S_i, A_i)
\label{longTime}
\end{equation}
is the sum of discounted future reward, $\gamma \in [0,1]$ is the discounted rate, and $r(S_i, A_i)$ is the reward at the state $S_i$ with action $A_i$ at time step $i$. (we will discuss the reward function later in Section \ref{stepFuniton}.)

\begin{figure}[htbp]
	\centering
	\includegraphics[width=0.48\textwidth]{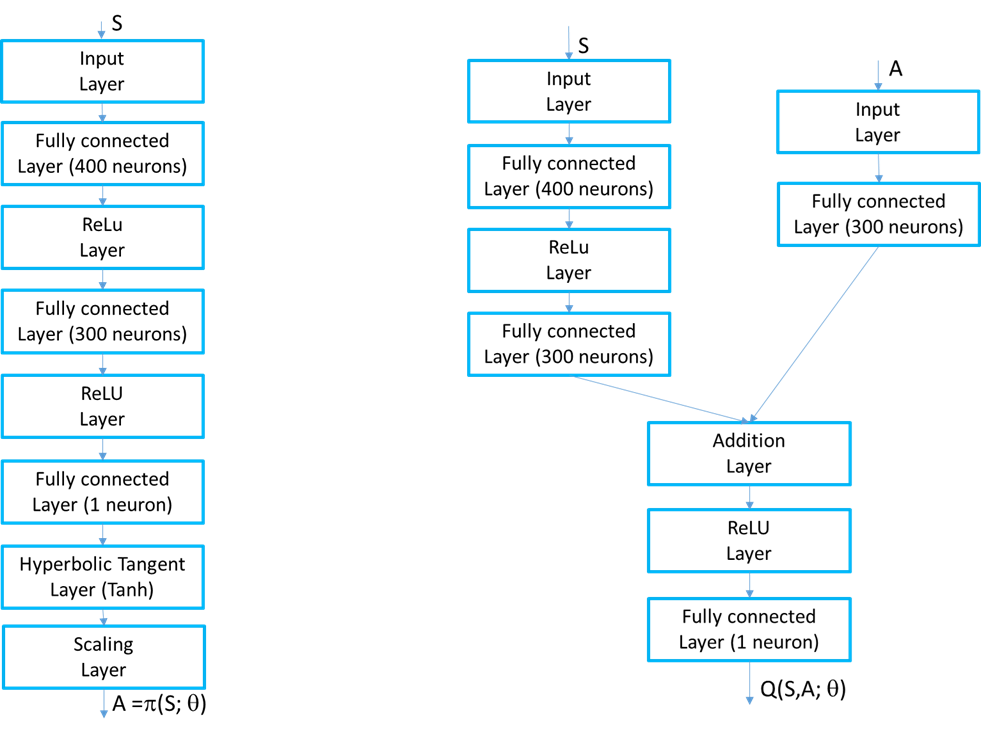}
	\caption{\label{Net} (a) Actor network (b) Critic network}
\end{figure}

Fig.~\ref{Net}(a) shows the actor network which takes input of a four-dimensional state (position coordinates x and y, agent previous heading, and time steps) and generates the action as its output (agent new heading). This neural network is designed the same as in \cite{lyw26} but we will use improved reward functions. Note that in DDPG, the action is proposed to add a stochastic Gaussian or Ornstein Uhlenbeck (OU) noise 
\begin{equation}
A_{\theta}=\mu(S;\theta_A)+N
\label{action}
\end{equation}
where the standard deviation of the noise $N$ is carefully selected and dependent on if the network is in exploration stage or exploitation stage. Large deviation allows the network to try out or explore new options in action space while the small deviation implies that the network is already good enough to propose an action therefore adding little change on the good action. Sections \ref{EandE} will discuss more on the implementations.

The critic network is also like the one used in \cite{lyw26} but a smart initial heading is used to avoid entering the ``no go'' zones in the early steps. Fig.~\ref{Net}(b) shows the critic network which uses a four-dimensional state and one-dimensional action as the input to estimate the Q value which is the expected long-term reward as defined in (\ref{cNetwork}) and (\ref{longTime}). At the beginning of each training episode, these networks will be reset (see more details in Section~\ref{reset}).

\subsection{Exploration and exploitation}\label{EandE}

Machine learning, in particular, reinforcement learning has two important features: exploration and exploitation. Exploration uses a $\epsilon$-greedy strategy to explore new actions for better solutions and exploitation utilizes the known best action to achieve the best performance. Our DDPG implementation achieves the balance by setting up a standard deviation $\sigma$ for the action noise in (\ref{action}) for the $\epsilon$-greedy strategy. It is important to select this parameter in the implementation. Our selection is $\sigma \in [3, 5]$ degrees in the first training process. When we get a satisfactory agent, we have an option to retrain the agent with a smaller $\sigma$ because we have more confidence with the trained agent.

\subsection{Modeling the environment: reset function}\label{reset}

In our implementation for the path planning problem. The reset function is invoked when a training episode starts. In the reset function, the agent restarts with the starting position and heading but keeps the recent network parameters. If the training episodes start with random starting positions, the starting positions will be generated randomly in reset function. When the training is completed and a test is performed, the starting position is provided by the test function and is passed to the reset function. 

The initial heading $\phi_0$ is set in reset function in both training and testing stages. A smart initial heading is proposed using the following simple heuristics: 

(a) If the line segment from starting point $P_0=(x_0,y_0)$ to the destination $P_f=(x_f,y_f)$ does not cross the restricted circular area(s), 
\begin{equation}
\phi_0 = \tan^{-1}\left( \frac{y_f-y_0}{x_f-x_0} \right).
\label{phi0a}
\end{equation}

(b) If the line segment from starting point $P_0$ to the destination $P_f$ does cross the restricted circular area(s), the two tangent points $P_1$ and $P_2$ from $P_0$ to the restricted circular area are calculated, then the distances from $P_f$ to $P_1$ and $P_2$ are calculated, denote the distances 
\begin{equation}
d(P_i,P_f)=\sqrt{(x_i-x_f)^2+(y_i-y_f)^2}, \hspace{0.1in} i=1,2,
\end{equation}
and 
\begin{equation}
P^*=\{ (x^*,y^*)~|~ \min (d(P_i,P_f)) \},
\end{equation}
the initial heading is given as 
\begin{equation}
\phi_0 = \tan^{-1}\left( \frac{y^*-y_0}{x^*-x_0} \right).
\label{phi0b}
\end{equation}

\begin{figure}[htbp]
\centering
\includegraphics[width=0.48\textwidth]{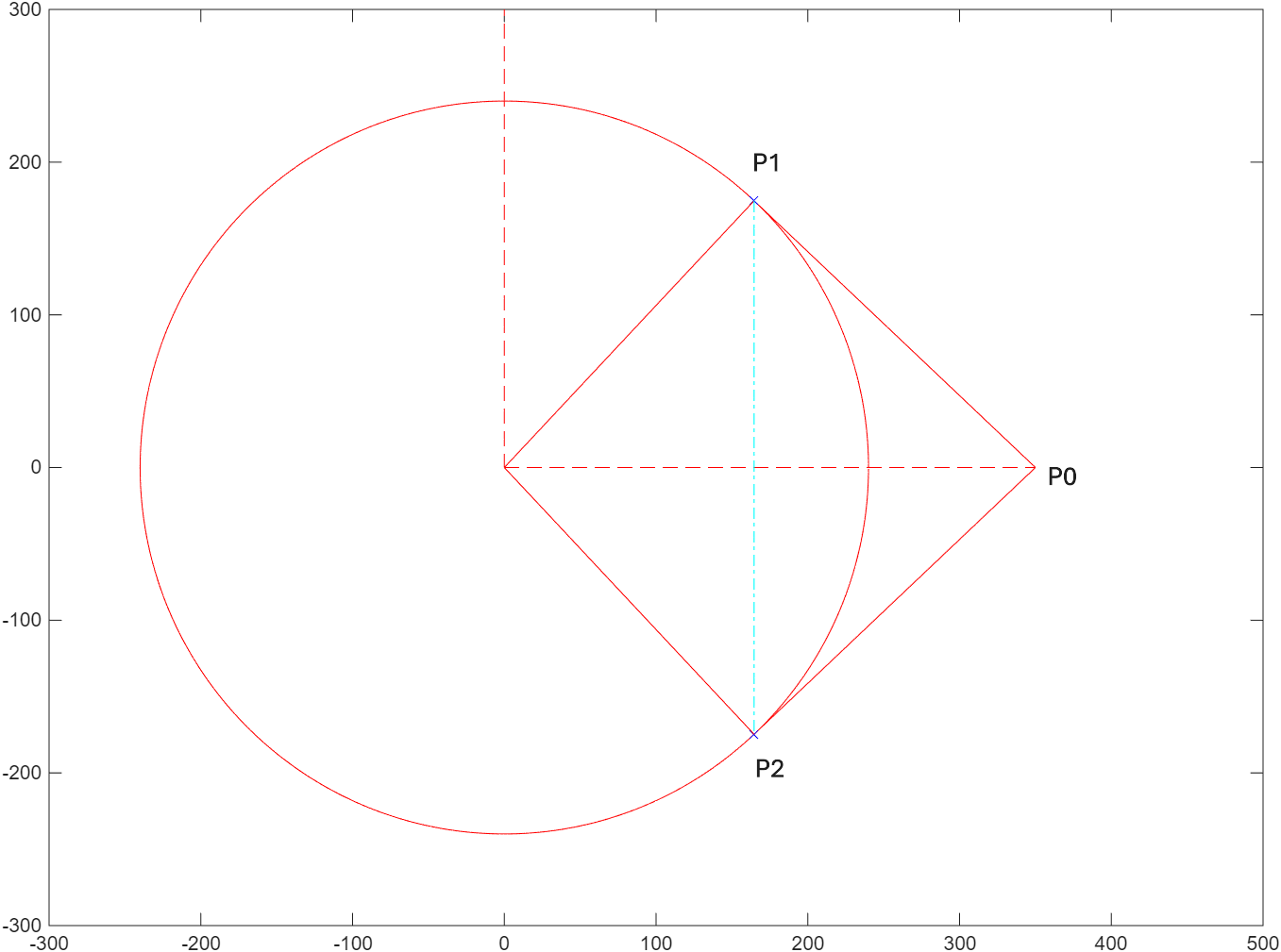}
\caption{\label{f:b1} The tangent points in a special case.}
\end{figure}

Therefore, the key is to calculate the tangent points $P_1$ and $P_2$. First, assuming that the obstacle is circular with the center at $(0,0)$, its radius is $r$ and $P_0=(x_0,0)$ is on the $x$-axis (see Fig.~\ref{f:b1}), it can easily derive the tangent points $P_1$ and $P_2$ are given by (\ref{SpecialTangetPoints}).

\begin{eqnarray}
\left[ \begin{array}{c} x_{1,2} \\ y_{1,2} \end{array} \right]
=\left[ \begin{array}{c} \frac{r^2}{x_0} \\ \pm \frac{r}{x_0} \sqrt{x_0^2-r^2} \end{array} \right].
\label{SpecialTangetPoints}
\end{eqnarray}

\begin{figure}[htbp]
\centering
\includegraphics[width=0.48\textwidth]{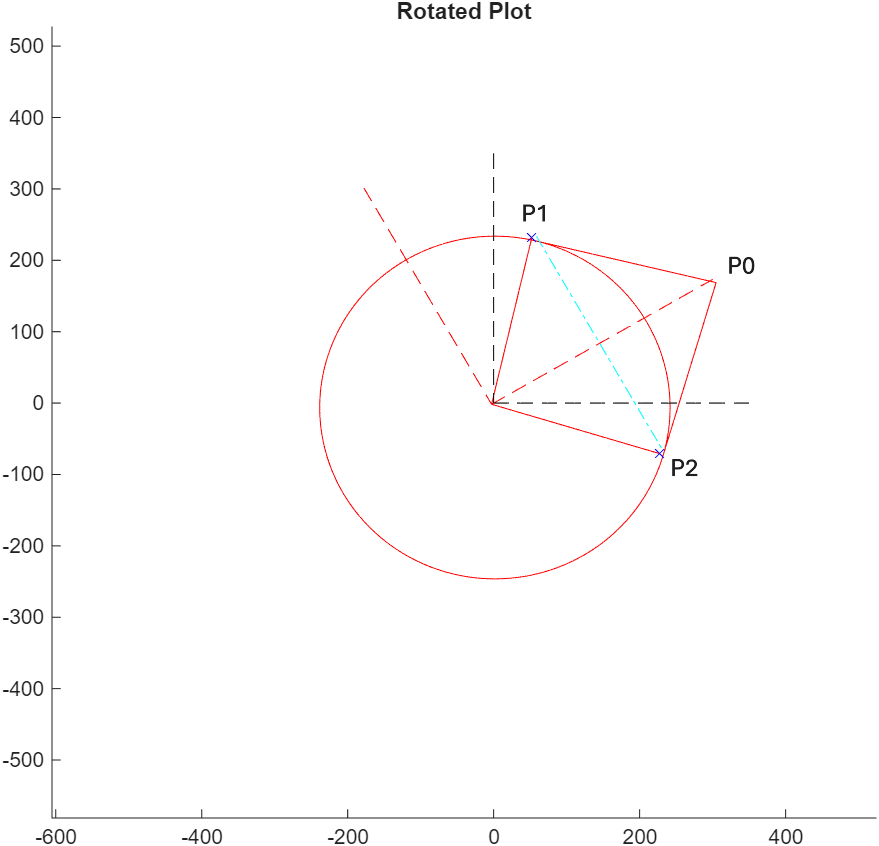}
\caption{\label{f:b2} The tangent points in a less special case.}
\end{figure}

If $P_0=(x_0,y_0)$ is not on the $x$-axis (see Fig.~\ref{f:b2}), then, a rotation transform shows that the tangent points are given by (\ref{tangetPoints}).

\begin{eqnarray}
\left[ \begin{array}{c} x_{1,2} \\ y_{1,2} \end{array} \right]
=\frac{r^2}{d_0^2} \left[ \begin{array}{c} x_{0} \\ y_{0} \end{array} \right] 
\pm \frac{r}{d_0^2} \sqrt{d_0^2-r^2} \left[ \begin{array}{c} - y_{0} \\ x_{0} \end{array} \right].
\label{tangetPoints}
\end{eqnarray}

\begin{figure}[htbp]
\centering
\includegraphics[width=0.48\textwidth]{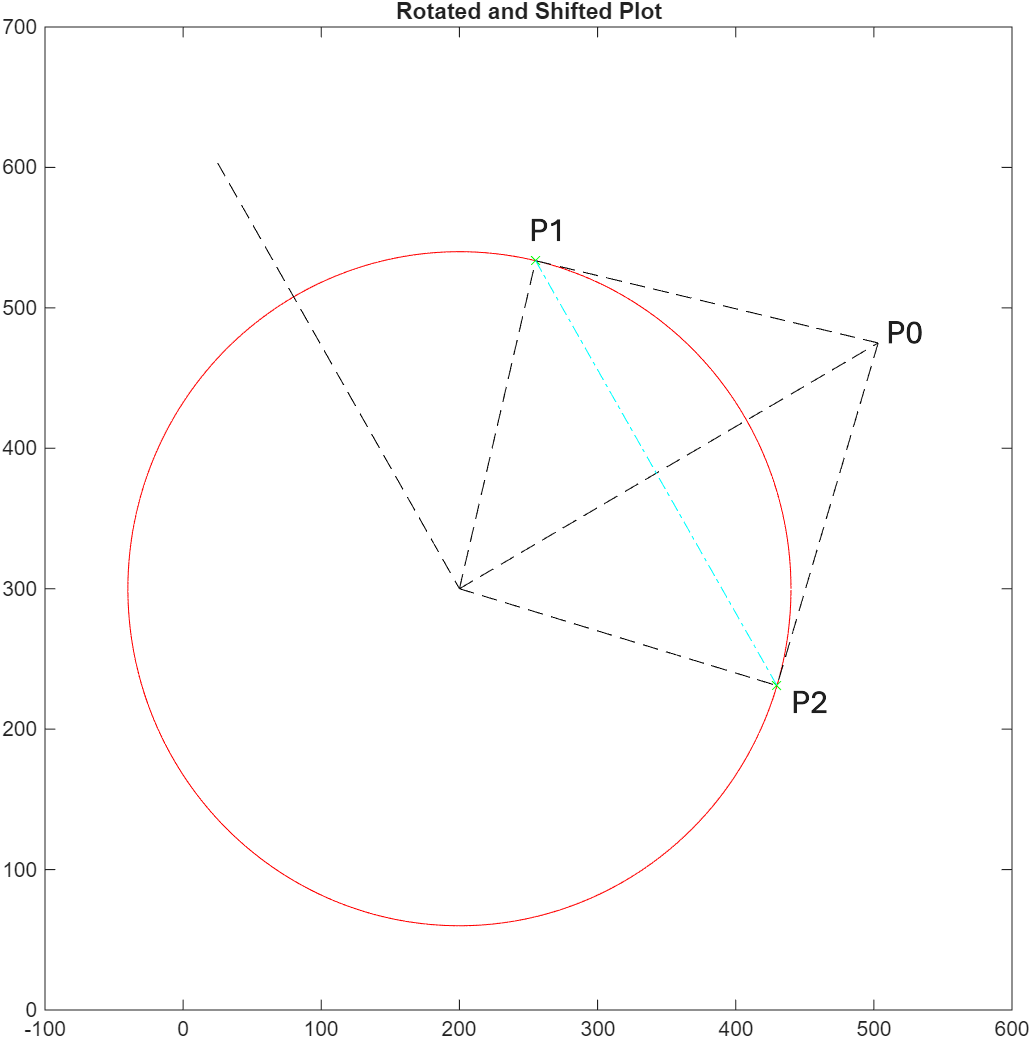}
\caption{\label{f:b3} The tangent points in general case.}
\end{figure}

If the center at $(x_c,y_c) \neq (0,0)$ is not at origin, and $P_0=(x_0,y_0)$ is not on the $x$-axis (see Fig.~\ref{f:b3}), then, a rotation and a shift transforms show that the tangent points are given by (\ref{generalTangetPoints}).

\begin{eqnarray}
\left[ \begin{array}{c} x_{1,2} \\ y_{1,2} \end{array} \right]
& =& \left[ \begin{array}{c} x_c \\ y_c \end{array} \right] + \frac{r^2}{d_0^2} \left[ \begin{array}{c} x_{0}-x_c \\ y_{0}-y_c \end{array} \right] 
\nonumber \\
& & 
\pm \frac{r}{d_0^2} \sqrt{d_0^2-r^2} \left[ \begin{array}{c} - y_{0}+y_c \\ x_{0}-x_c \end{array} \right]. 
\label{generalTangetPoints}
\end{eqnarray}

This general formula is used in multiple obstacle scenarios to be discussed in Section~\ref{multiObs}.

\subsection{Modeling the environment: step function}\label{stepFuniton}

While the reset function is invoked when a new training episode starts, the step function will be used repeatedly in every step in both training and test stages. It is important to keep in mind that the step function used in both training and test stages must be the same.

A critical part of using DDPG machine learning method is to set an appropriate reward function, which is accomplished in step function. Using artificial potential fields (PF) as part of the reward function for path planning is a widely adopted strategy to guide the agent to approach the destination and prevent the agent from entering the block-out zone \cite{yao2020path,zmhzy23,pjzwys23}, which was probably proposed first by Hogan in \cite{hogan1984} (DDPG didn't exist at that time). We also use artificial potential fields (PF) as part of our reward function. Let $P_t=[x_t,y_t]^T$ be the current agent location. The update of the state is given by: 
\begin{equation}
\phi_{t+1} = \phi_t +|A_{t}|,
\label{phi}
\end{equation}
and
\begin{equation}
P_{t+1} = P_{t} + |V|*[\cos(\phi_{t+1}),\sin(\phi_{t+1})]^T*T_s,
\label{xUpdated}
\end{equation}
where $T_s=0.25$ second is the time step, and $P_{t+1}$ is the agent location at the next time step. Let the center of the obstacle is at $P_c=[0,0]^T$. Let $N(P_f,P_f,\sigma_1)$ be the value at $P_f$ of the Gaussian distribution with mean $P_f$, and $\sigma_1$ as its standard deviation, and $N(P_t,P_f,\sigma_1)$ be the value at $P_t$ of this Gaussian distribution with mean $P_f$, and variance $\sigma_1$ as its standard deviation, and the normalized Gaussian distributions are given by:  
\begin{equation}
\tilde{N}(P_t,P_f,\sigma_1)=\frac{N(P_t,P_f,\sigma_1)}{N(P_f,P_f,\sigma_1)}
\label{normalied}
\end{equation}
\begin{equation}
\tilde{N}(P_{t+1},P_f,\sigma_1)=\frac{N(P_{t+1},P_f,\sigma_1)}{N(P_f,P_f,\sigma_1)}
\label{normalied1}
\end{equation}
where 
\begin{equation}
\sigma_1 = \left[ \begin{array}{cc}
57600 & 100 \\ 100 & 57600 
\end{array} \right].
\label{sigma1}
\end{equation}
The reward for the agent moving toward the destination is modeled by an attractive artificial potential field which is represented by the difference of the normalized Gaussian distributions:
\begin{equation}
\tilde{N}(P_{t+1},P_f,\sigma_1) - \tilde{N}(P_{t},P_f,\sigma_1).
\end{equation}

The penalty of entering the restricted area is modeled in a similar way by a repulsive artificial potential field which is represented by a difference of two normalized Gaussian distributions: 
\begin{equation}
\tilde{N}(P_t,P_c,\sigma_2)=\frac{N(P_t,P_c,\sigma_2)}{N(P_c,P_c,\sigma_2)}
\label{NPc}
\end{equation}
\begin{equation}
\tilde{N}(P_{t+1},P_c,\sigma_2)=\frac{N(P_{t+1},P_c,\sigma_2)}{N(P_c,P_c,\sigma_2)}
\label{NPc1}
\end{equation}
where 
\begin{equation}
\sigma_2= \left[ \begin{array}{cc}
57100 & 100 \\ 100 & 57100 
\end{array} \right].
\label{sigma2}
\end{equation}

The penalty of energy consumption is modeled by $|A|^2$, which implicitly minimizes the path length. 
The combined reward function is given as

\begin{eqnarray}
r(P_t,A_t) = w_1 \left[ \tilde{N}(P_{t+1},P_f,\sigma_1) - \tilde{N}(P_{t},P_f,\sigma_1) \right] \nonumber \\
- w_2  \left[ \tilde{N}(P_{t+1},P_c,\sigma_2) - \tilde{N}(P_t,P_c,\sigma_2) \right] \nonumber \\
- w_3 |A_t|^2 -10,
\label{e:reward}
\end{eqnarray}
where $w_1$, $w_2$, and $w_3$ are carefully selected weights to be discussed below; for each step that does not reach the destination, we add a discrete penalty $-10$ in the reward function to minimize the number of time steps per episode of training so that the travel time of the agent could be minimized.

\section{Path planning with a single obstacle}\label{singleObs}

In this section, we consider a simple path planning problem that is the same as in \cite{lyw26}. We explain how training and re-training can be used to balance exploration and exploitation. We will show that the inclusion of a smart initial heading and a reward for moving toward the destination in every step proposed in this paper does significantly improve the performance of the trained agent. All environment parameters are the same as in \cite{lyw26} for the comparison purpose. The starting position can be any place in a square of size $500 \times 500$ but outside of the restricted zone whose center is located at $(0,0)$ and its radius is $240$, the desired destination is at $(-200,-400)$.

\subsection{Heat map of a single obstacle}

\begin{figure}[htbp]
\centering
\includegraphics[width=0.48\textwidth]{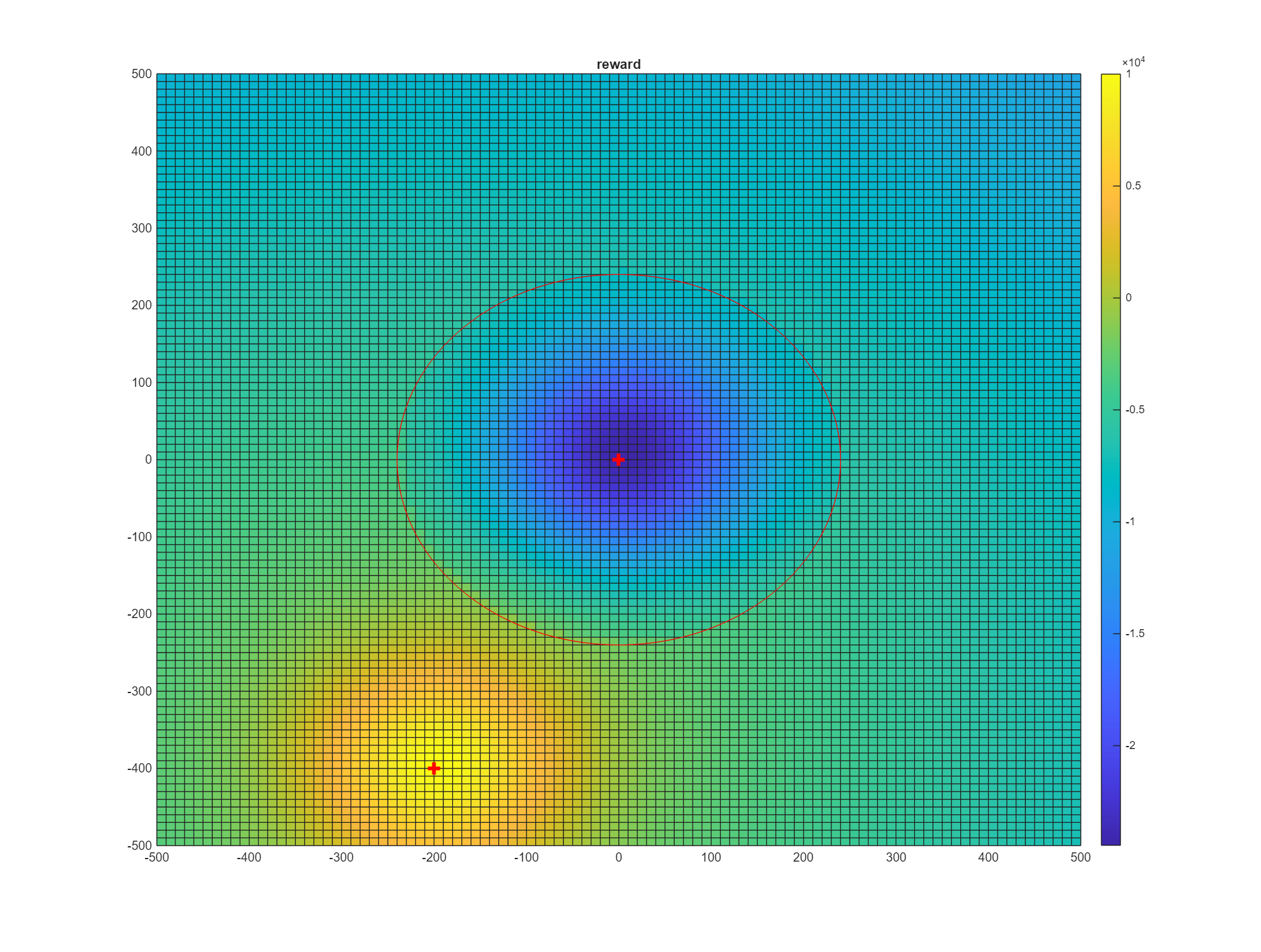}
\caption{\label{f:p1} The heat map of the reward function in (\ref{e:reward2}) where $w_3=0$}
\end{figure}

Similar to the idea used in \cite{yzqygzly20}, a heat map that represents artificial attractive potential field and artificial repulsive potential field is used for assisting the selection of the weights of $w_1$, $w_2$, and $w_3$. The heat map is as described in Fig.~\ref{f:p1}. The weights $w_1=5000$, $w_2=10$, and $w_3=-0.1$ are carefully selected so that going around the restricted zone has reward advantages than crossing it. In addition, the large weight on the attractive reward aims for the agent to find the goal destination fast, and the small weight on the action allows for the possibilities of sudden or large change in heading.

%


We trained the DDPG agent with large standard deviation in the exploration stage to reach a certain level of reward threshold. Then we lowered the standard deviation in the exploitation stage to refine the agent to maximize the above accumulated reward.

%

%

\subsection{Effects of smart initial heading and variations of reward}

In this research, we focus on the capabilities of the trained agent. If the agent starts with an initial position where it is never trained before, would it be able to avoid the obstacles and find the destination? We define a feasible set of points which the agent can start from to find a feasible path to the desired destination without crossing the obstacle. We would also like to know what the benefit is for introducing a smart initial heading (\ref{generalTangetPoints}) and variations of the reward defined in (\ref{e:reward}) which were not used in \cite{lyw26}.

\begin{figure}[htbp]
\centering
\includegraphics[width=0.48\textwidth]{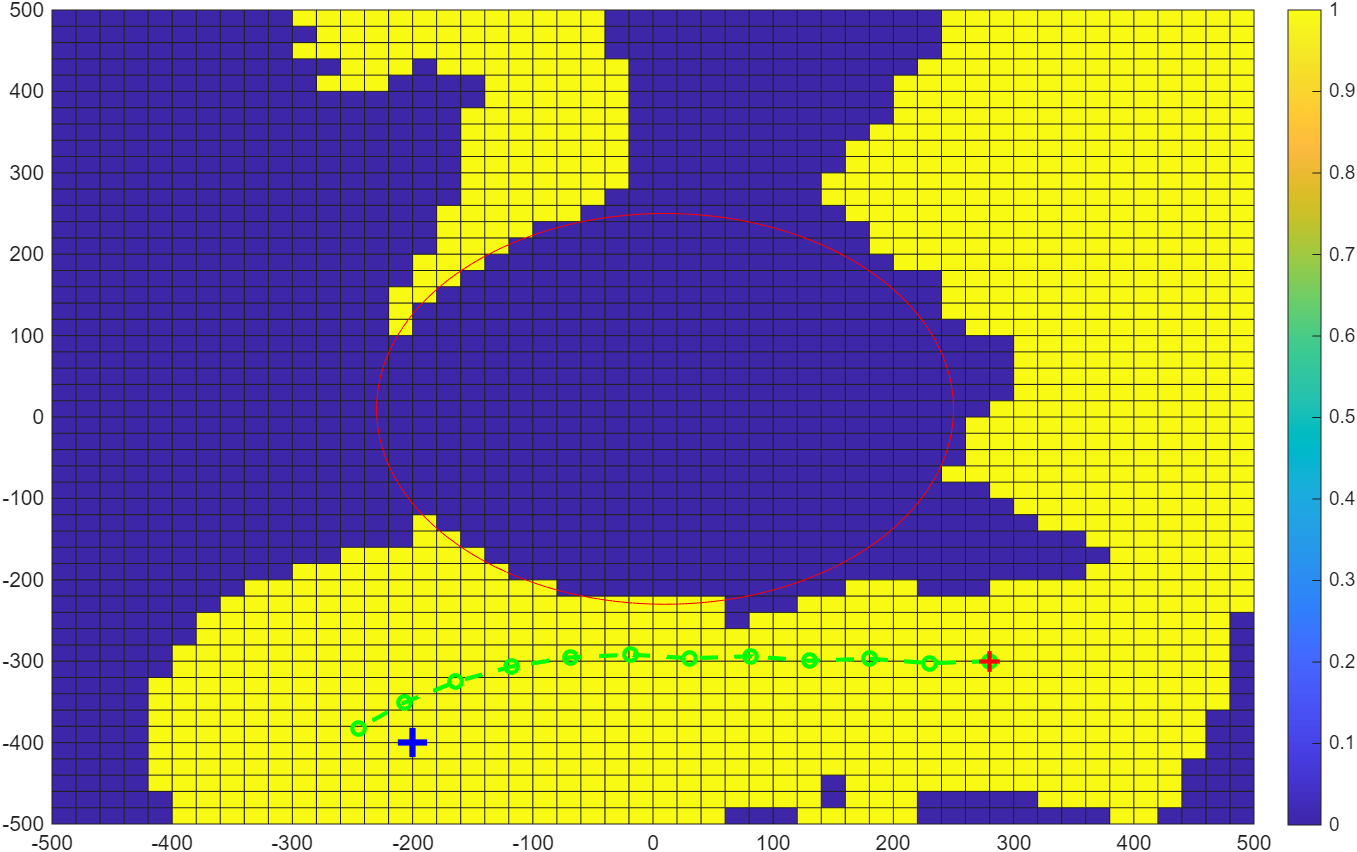}
\caption{\label{f:feasibleArea1} The feasible set with a trained agent using random starting points, but no the smart initial heading and trained with the reward in (~\ref{e:reward2}).}
\end{figure}

\begin{figure}[htbp]
\centering
\includegraphics[width=0.6\textwidth]{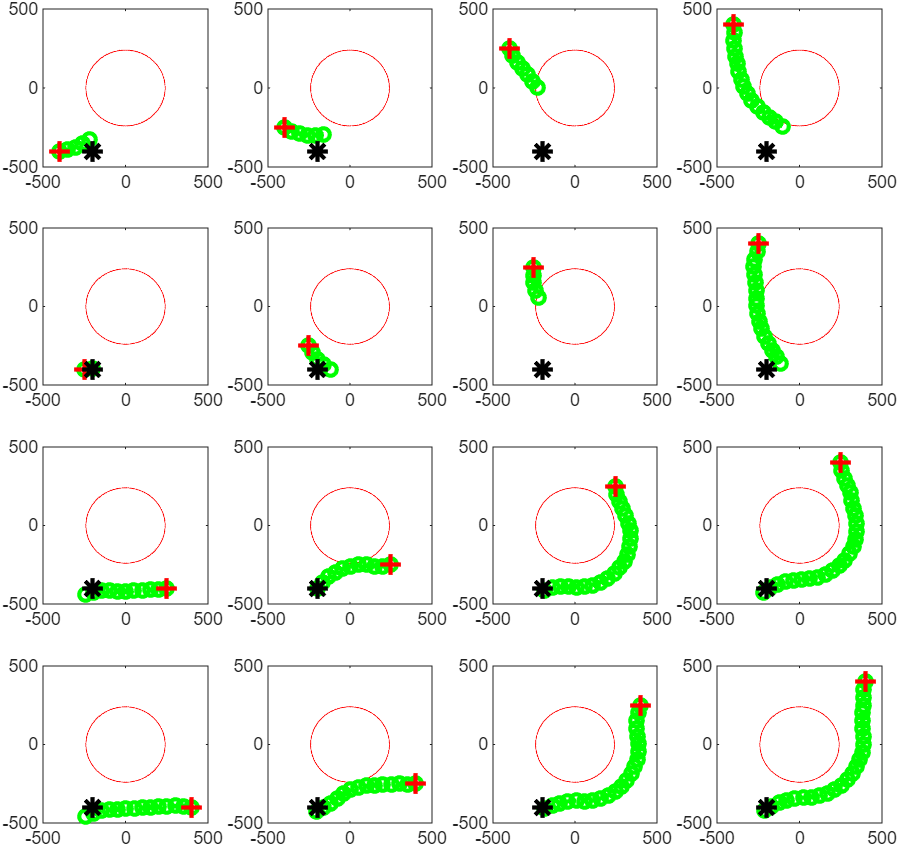}
\caption{\label{f:path_no_heading} Paths obtained from agent that does not use smart initial heading}
\end{figure}

\begin{figure}[htbp]
\centering
\includegraphics[width=0.48\textwidth]{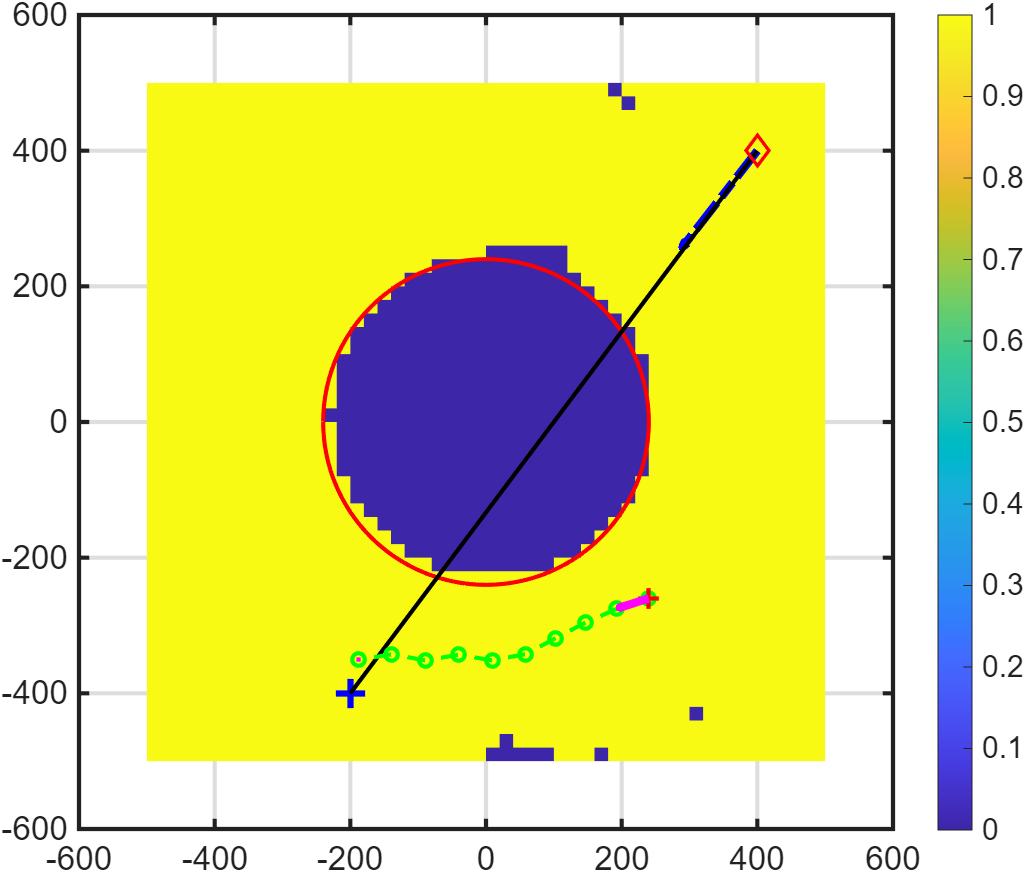}
\caption{\label{f:feasibleArea2} The feasible set with a trained agent using random starting points, the smart initial heading, and trained with the reward in (~\ref{e:reward})}
\end{figure}

\begin{figure}[htbp]
	\centering
	\includegraphics[width=0.6\textwidth]{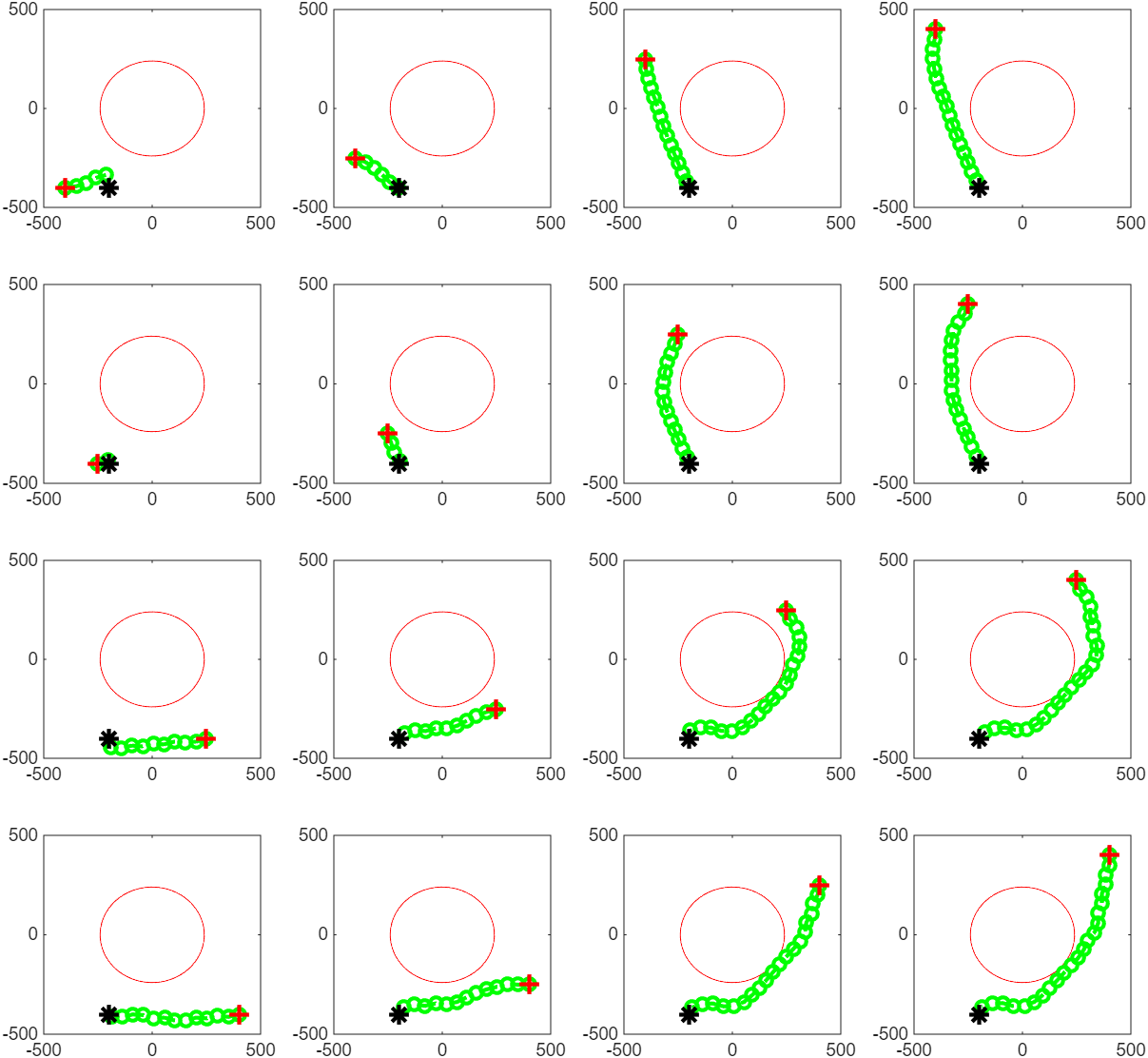}
	\caption{\label{f:paths_smart_heading} Paths obtained from agent that uses smart initial heading}
\end{figure}

To achieve this goal, the starting points are randomly selected in every episode in the training process. After the training is finished, we test the trained agent using starting points equally distributed in the grid and check if the agent can find a feasible path that does not enter the restricted area and does approach the destination. The feasible set of a trained agent {\it without} using the smart initial heading and with the reward function in(~\ref{e:reward2}) is represented in Fig.~\ref{f:feasibleArea1}. We also examined paths inside and outside of the feasible set to verify the feasible set is correctly identified. Examples of these paths are displayed in Fig.~\ref{f:path_no_heading}. The improved feasible set of a trained agent {\it with} the smart initial heading and the reward accounting for the reward gain in (~\ref{e:reward}) is represented in Fig.~\ref{f:feasibleArea2}. We also examined paths inside and outside of the feasible set to verify the feasible set is correctly identified. Examples of these paths are displayed in Fig.~\ref{f:paths_smart_heading}. Comparing Fig.~\ref{f:feasibleArea1} and Fig.~\ref{f:feasibleArea2}, it is obvious that introducing the smart initial heading and the reward of moving toward the destination improves significantly the agent performance.

\begin{eqnarray}
	r(P_t,A_t) = w_1 \left[\tilde{N}(P_{t},P_f,\sigma_1) \right] \nonumber \\
	- w_2  \left[\tilde{N}(P_t,P_c,\sigma_2) \right] \nonumber \\
	- w_3 |A_t|^2,
	\label{e:reward2}
\end{eqnarray}

\section{Path planning with multiple obstacles}\label{multiObs}

\begin{figure}[htbp]
\centering
\includegraphics[width=0.98\textwidth]{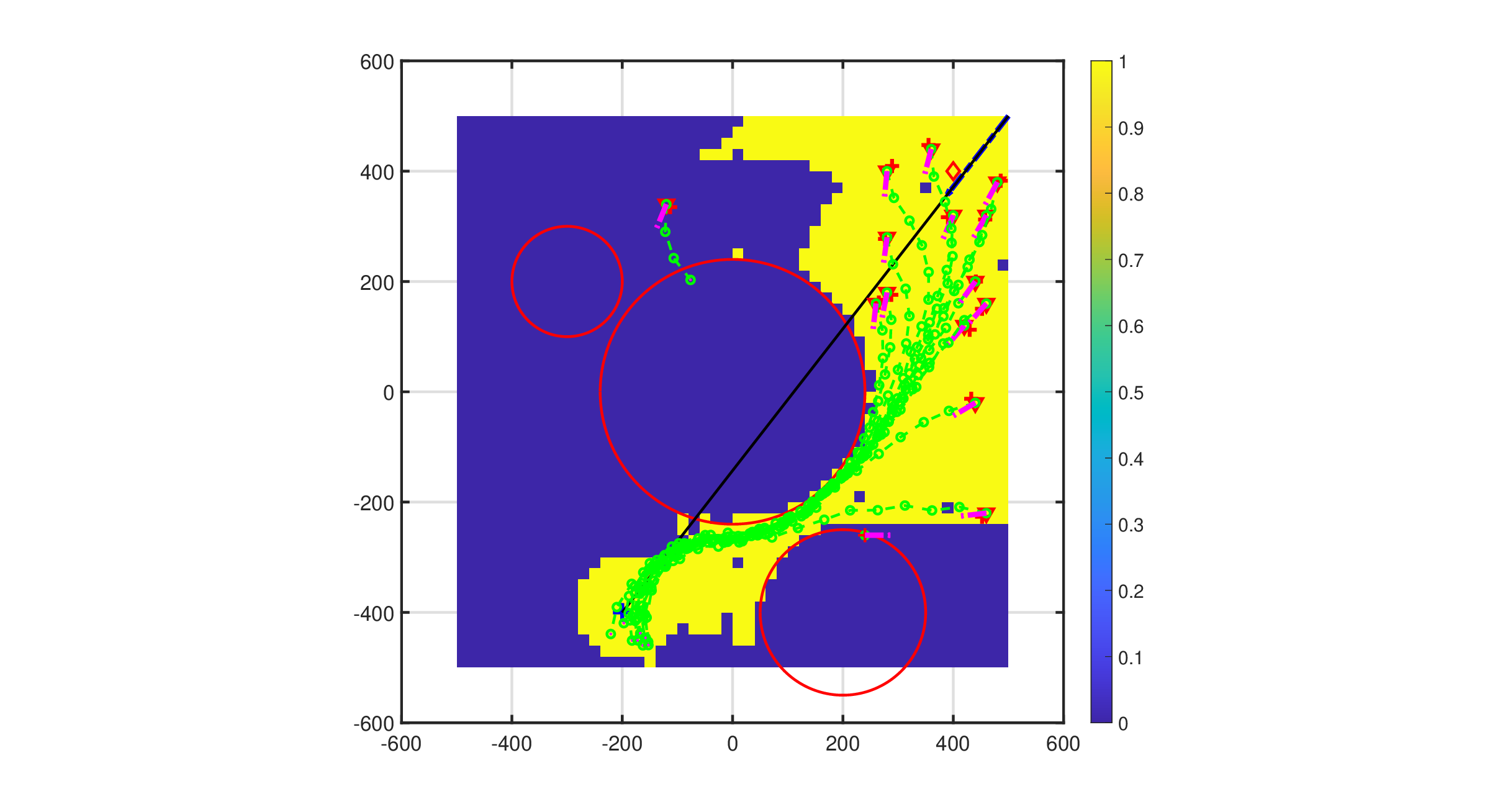}
\caption{\label{f:training4} Feasible area of the specialized agent who is trained using (400, 400) in 3 obstacle scenario, where testing examples are in green}
\end{figure}

In this section, we consider a complicated scenario where there are three obstacles, as shown in \ref{f:training4} . Their centers $P_i$ are at $(0,0)$, $(200,-400)$, $(-300, 200)$, and their radius are $240$, $150$, and $100$ respectively. The desired destination is at $(-200,-400)$. The goal is to train a specialized agent from a starting point $(400 400)$ through the narrow passage between the bottom blocks to reach the destination.  

We use the same actor and critic neural networks as the ones in Section \ref{singleObs} and \ref{DDPGmethod}. The reward function (~\ref{e:reward1} is modified using trial and error approach to generate the best performance in our best knowledge, where the weights are dominated by the attraction gain $w_1=5000,w_2=120, w_3=0.1$. 
\begin{eqnarray}
	r(P_t,A_t) = w_1 \left[ \tilde{N}(P_{t+1},P_f,\sigma_1) - \tilde{N}(P_{t},P_f,\sigma_1) \right] \nonumber \\
	- w_2  \sum_{i=1}^{i=3}\tilde{N}(P_{t+1},P_i,\sigma_2) \nonumber \\
	- w_3 |A_t|^2 -10,
	\label{e:reward1}
\end{eqnarray}
The covariance matrix for the artificial fields are 
\begin{equation}
	\sigma_1=\sigma_2= \left[ \begin{array}{cc}
		40000 & 100 \\ 100 & 40000 
	\end{array} \right].
	\label{e:Msigma}
\end{equation}

Fig.~\ref{f:training4} shows the feasible area of the specialized agent to avoid the multiple obstacle case where the initial smart heading is denoted in pink using a general formula (\ref{generalTangetPoints}). The feasible area of the specialized agent is the neighboring area of the optimal path from the special initial point to the destination. This observation justifies the “principle of optimality” for using the Bellman equation in DDPG, which states that an optimal policy contains optimal sub-policies.
The examples of resulting paths when testing the specialized
agent are marked as green. Good paths are obtained when the
starting positions are within the proximity of the specialized
trained position.

%

\section{Conclusion and Future Work}\label{Conc}

In this paper, we have developed a DDPG-based reinforcement method for path planning problems in the presence of obstacles. Mablab code is implemented and simulations for problems with single and triple obstacles are performed. 
We will consider other methods for the path planning problem in the future, for example, we will investigate if combining machine learning with some optimization method \cite{yang25} is an attractive alternative.

\section*{Acknowledgments}

This material is based on work supported by the Department of the Air Force under the 
Air Force Contract No.FA955023D0001. Any opinions, findings, conclusions, or recommendations expressed 
in this material are those of the author(s) and do not necessarily reflect the views of the Department of the Air 
Force.


\begin{thebibliography}{00}
\bibitem{hctch18} X. Hu, L. Chen,  B. Tang,  D. Cao, and H. He, Dynamic path planning for autonomous driving on various roads with avoidance of static and moving obstacles,  Mechanical Systems and Signal Processing, 100(1), pp. 482-500,  2018.
\bibitem{msml11} J. Mason, J. Stupl, W. Marshall, and C. Levit, Orbital debris–debris collision avoidance, Advances in Space Research, 48(10), pp. 1643-1655, 2011.
\bibitem{dzwv23} P. M. Dillon, M. D. Zollars, I. E. Weintraub, and A. Von Moll, Optimal trajectories for aircraft avoidance of multiple weapon engagement zones, Journal of Aerospace Information Systems, 2023.
\bibitem{vw24} I. A. Von Moll, and Weintraub, Basic engagement zones. Journal of Aerospace Information Systems, 21(10), pp.885-891, 2024.
\bibitem{wvchf22} I.E. Weintraub, and A. Von Moll, C.A. Carrizales, N. Hanlon, and Z.E. Fuchs, An optimal engagement zone avoidance scenario in 2-D, In AIAA SciTech 2022 Forum (p. 1587), 2022.
\bibitem{gc13} E. Galceran, and M. Carreras, A survey on coverage path planning for robotics, Robotics and Autonomous systems, 61(12):1258-76, 2013.
\bibitem{ask15} Z. Algfoor, M. S. Sunar, and H. Kolivand, A Comprehensive study on pathfinding techniques for robotics and video games, International Journal of Computer Games Technology, Vol. 2015, Article ID 736138, 2015.
\bibitem{bg07} P. Bhattacharya and M. L. Gavrilova, Voronoi diagram in optimal path planning, 4th International Symposium on Voronoi Diagrams in Science and Engineering, 2007.
\bibitem{zwlw22} X. Zheng, Z. Wang, D. Liu, and H. Wang, A path planning algorithm for PCB surface quality automatic inspection, Journal of Intelligent Manufacturing 33(6), pp. 1829-1841, 2022.
\bibitem{up23}  D. Utyamishev and I. Partin-Vaisband, Multiterminal pathfinding in practical VLSI systems with deep neural networks, ACM Transactions on Design Automation of Electronic Systems, 28(4), Article 51, 2023.
\bibitem{herber15} D. R. Herber, Basic implementation of multi-interval pseudospectral methods to solve optimal control problem, UIUC technical report, UIUC-ESDL-2015-01, 2015.
\bibitem{hwmz20} L. Hewing, K. P. Wabersich, M. Menner, and M. N. Zeilinger, Learning-based model predictive control: toward safe learning in control, Annual Review of Control, Robotics, and Autonomous Systems, 3, 269–96, 2020.
\bibitem{nstj19} H. Niu, A. Savvaris, A. Tsourdos, and Ze Ji, Voronoi-visibility roadmap-based path planning algorithm for unmanned surface vehicles, The Journal of Navigation, 72 (4), pp. 850–874, 2019.
\bibitem{watkins89} C. J. C. H. Watkins, Learning from delayed rewards. Ph.D. thesis, Cambridge University,
1989.
\bibitem{cx18} Y. Chao, and X. Xiang, A path planning algorithm for UAV based on improved Q-learning,
In 2018 2nd international conference on robotics and automation sciences (ICRAS), pp. 1-5.
IEEE, 2018.
\bibitem{mh20} A. Maoudj, and A. Hentout, Optimal path planning approach based on Q-learning algorithm
for mobile robots, Applied Soft Computing, 97, 106796, 2020.
\bibitem{prppl24} A. Puente-Castro, D. Rivero, E. Pedrosa, A. Pereira, A. Lau, \& E. Fernandez-Blanco, Q-learning based system for path planning with unmanned aerial vehicles swarms in obstacle environments. Expert Systems with Applications, 235, 121240, 2024.
\bibitem{syc23} A. Sonny, S. R. Yeduri, and L. R. Cenkeramaddi, Q-learning-based unmanned aerial vehicle
path planning with dynamic obstacle avoidance. Applied Soft Computing, 147, 110773, 2023.
\bibitem{wyl22} C. Wang, X. Yang, and H. Li, Improved q-learning applied to dynamic obstacle avoidance
and path planning, IEEE Access 10, pp. 92879-92888, 2022.
\bibitem{zzzy17} Y. Zhao, Z. Zheng, X. Zhang, and L. Yang, Q learning algorithm based UAV path learning and
obstacle avoidence approach, In 2017 36th Chinese control conference (CCC), pp. 3397-3402, IEEE, 2017.
\bibitem{mksgawr13} V. Mnih, K. Kavukcuoglu, D. Silver, A. Graves, I. Antonoglou, D. Wierstra, and M. Riedmiller, Playing Atari with Deep Reinforcement Learning, ArXiv:1312.5602 [Cs], December 19, 2013. 
\bibitem{mnih15} V. Mnih,  et al., Human-level control through deep reinforcement learning, Nature 518(7540), pp. 529-533, 2015.
\bibitem{mbmglhsk16} V. Mnih, A. P. Badia, M. Mirza, A. Graves, T. Lillicrap, T. Harley, D. Silver, and K. Kavukcuoglu, Asynchronous methods for deep reinforcement learning. In International Conference on Machine Learning, pp. 1928–1937, 2016.
\bibitem{cpll22} P. Chen, J. Pei, W. Lu, and M. Li, A deep reinforcement learning based method for real-time
path planning and dynamic obstacle avoidance, Neurocomputing 497 pp. 64-75, 2022.
\bibitem{gzlsx23} Y. Gu, Z. Zhu, J. Lv, L. Shi, Z. Hou, and S. Xu, DM-DQN: Dueling Munchausen deep Q network for robot path planning, Complex \& Intelligent Systems 9(4), pp. 4287-4300, 2023.
\bibitem{hza25} Le Han, H. Zhang, and N. An, A continuous space path planning method for unmanned aerial vehicle based on particle swarm optimization-enhanced deep q-network,  Drones 9(2),  122, 2025.
\bibitem{hqll23} R. Huang, C. Qin, J. Li, and X. Lan, Path planning of mobile robot in unknown dynamic continuous environment using reward‐modified deep Q‐network, Optimal Control Applications and Methods, 44(3) pp. 1570-1587, 2023.
\bibitem{lzd18} X. Lei,  Z. Zhang, and P. Dong, Dynamic path planning of unknown environment based on deep reinforcement learning, Journal of Robotics 2018(1), 5781591, 2018.
\bibitem{nkm23} T. Nakamura,  M. Kobayashi, and N. Motoi, Path planning for mobile robot considering turnabouts on narrow road by deep Q-network, IEEE Access 11, pp. 19111-19121, 2023.
\bibitem{wzdlzll23} W. Wang, G. Zhang, Q. Da, D. Lu, Y. Zhao, S. Li, and D. Lang,  Multiple unmanned aerial vehicle autonomous path planning algorithm based on whale-inspired deep Q-network, Drones 7(9), 572, 2023.
\bibitem{xyj25}  T. Xie, , X. Yao, Z. Jiang, et al. AGV path planning with dynamic obstacles based on deep Q-network and distributed training. Int. J. of Precis. Eng. and Manuf.-Green Tech. 12, 1005–1021 (2025).
\bibitem{ylp20} Y. Yang, J, Li. and L, Peng, Multi-robot path planning based on a deep reinforcement learning DQN algorithm, CAAI Trans. Intell. Technol., 5,  177-183, 2020.
\bibitem{yzqygzly20} Q. Yao, Z. Zheng, L. Qi, H. Yuan, X. Guo, M. Zhao, Z. Liu, and T. Yang, Path planning method with improved artificial potential field—a reinforcement learning perspective, IEEE access 8, 135513-135523, 2020.
\bibitem{zlxg18}  S. Zhou, X. Liu, Y. Xu, and J. Guo, A deep Q-network (DQN) based path planning method for mobile robots, In 2018 IEEE International Conference on Information and Automation (ICIA), pp. 366-371. IEEE, 2018.
\bibitem{lhphetsw15} T.P. Lillicrap, J. J. Hunt, A. Pritzel, N. Heess, T. Erez, Y. Tassa, D. Silver, and D. Wierstra, Continuous control with deep reinforcement learning, arXiv preprint arXiv:1509.02971, 2015.
\bibitem{slhdwr14} D. Silver, G. Lever, N. Heess, T. Degris, D. Wierstra, M. Riedmiller, Deterministic policy gradient algorithms, Proceedings of the 31 st International Conference on Machine Learning, Beijing, China, 2014.
\bibitem{asz24} A. A. Ali,  J. Shi, and Z. H. Zhu, Path planning of 6-DOF free-floating space robotic manipulators using reinforcement learning, Acta Astronautica, 224 367-378, 2024.
\bibitem{akr23} K. Almazrouei, I. Kamel, and T. Rabie, Dynamic obstacle avoidance and path planning through reinforcement learning, Applied Sciences 13(14), 8174, 2023.
\bibitem{gylwc23} X. Gao, L. Yan, Z. Li, G. Wang, and I. Chen, Improved deep deterministic policy gradient for dynamic obstacle avoidance of mobile robot, IEEE Transactions on Systems, Man, and Cybernetics: Systems 53(6), 3675-3682, 2023.
\bibitem{haws20} L. He,  N. Aouf, J. F. Whidborne, and B. Song, Deep reinforcement learning based local planner for UAV obstacle avoidance using demonstration data, arXiv preprint arXiv:2008.02521, 2020.
\bibitem{rhv23} M. Ramezani, H. Habibi, and H. Voos, UAV path planning employing MPC-reinforcement learning method considering collision avoidance, In 2023 International Conference on Unmanned Aircraft Systems, ICUAS, Warsaw, Poland, 2023.
\bibitem{whlm23} S. Wang, Y.i Hu, Z. Liu, and L. Ma, Research on adaptive obstacle avoidance algorithm of robot based on DDPG-DWA, Computers and Electrical Engineering, 109, 108753,  2023.
\bibitem{wcwzh18} S. Wen, J. Chen, S. Wang, H. Zhang, and X. Hu, Path planning of humanoid arm based on deep deterministic policy gradient, In 2018 IEEE International Conference on Robotics and Biomimetics (ROBIO), pp. 1755-1760. IEEE, 2018.
\bibitem{xcayz22} X. Xu, P. Cai, Z. Ahmed, V. S. Yellapu, and W. Zhang, Path planning and dynamic collision avoidance algorithm under COLREGs via deep reinforcement learning, Neurocomputing 468 (2022): 181-197.
\bibitem{yao2020path} Q. Yao, Z. Zheng, L. Qi, H. Yuan, X. Guo, M. Zhao, Z. Liu, and T. Yang, Path planning method with improved artificial potential field—a reinforcement learning perspective, IEEE access 8, 135513-135523, 2020.
\bibitem{zmhzy23} T. Zhu, J. Mao, L. Han, C. Zhang, and J. Yang, Real-time dynamic obstacle avoidance for robot manipulators based on cascaded nonlinear MPC with artificial potential field, IEEE Transactions on Industrial Electronics, 71(7), 7424-7434, 2023.
\bibitem{pjzwys23} R. Pan, L. Jie, X. Zhao, H. Wang, J. Yang, and J. Song, Active obstacle avoidance trajectory planning for vehicles based on obstacle potential field and MPC in V2P scenario, Sensors 23(6), 3248, 2023.
\bibitem{hogan1984} N. Hogan, Impedance control: An approach to manipulation, American control conference, 304-313, 1984.
\bibitem{lyw26} Q. Le, Y. Yang, and I. Weintraub, A Comparison of Reinforcement Learning and Optimal Control Methods for Path Planning, AAAI 2026 Spring Symposium Series, Burlingame, CA, USA, April 7-9, 2026.
\bibitem{yang25} Y. Yang, An arc-search interior-point algorithm for nonlinear constrained optimization, Computational Optimization and Applications, 90(3), (2025), 969-995.
\end{thebibliography}
\end{document}